\documentclass[10pt,twocolumn,letterpaper]{article}

\usepackage{cvpr}              %

\usepackage{graphicx}
\usepackage{amsmath}
\usepackage{amssymb}
\usepackage{booktabs}

\usepackage{multirow}
\usepackage{enumerate}
\usepackage{color}
\usepackage{xspace}
\usepackage{booktabs}
\usepackage{makecell}
\providecommand{\etal}{\textit{et al.}\xspace}
\providecommand{\mypara}[1]{{\noindent{\bf #1}}}
\usepackage{enumitem}

\newcommand\blfootnote[1]{%
  \begingroup
  \renewcommand\thefootnote{}\footnote{#1}%
  \addtocounter{footnote}{-1}%
  \endgroup
}

\usepackage[pagebackref,breaklinks,colorlinks]{hyperref}

\newcommand{\methodName}{\textsc{Phone2Proc}\xspace}
\newcommand{\procthor}{\textsc{ProcTHOR}\xspace}
\newcommand{\onesimlong}{Reconstructed Simulation\xspace}
\newcommand{\onesim}{\textsc{Recon}\xspace}

\usepackage[capitalize]{cleveref}
\crefname{section}{Sec.}{Secs.}
\Crefname{section}{Section}{Sections}
\Crefname{table}{Table}{Tables}
\crefname{table}{Tab.}{Tabs.}

\def\cvprPaperID{3485} %
\def\confName{CVPR}
\def\confYear{2023}

\usepackage{xcolor}
\definecolor{myblue}{HTML}{226DA3}
\hypersetup{
    citecolor=myblue
}

\begin{document}

\title{Phone2Proc: Bringing Robust Robots Into Our Chaotic World}

\author{
\\[-0.3in]\textbf{Matt Deitke$^{*\dagger\psi}$, Rose Hendrix$^{*\dagger}$, Luca Weihs$^\dagger$}\\\textbf{Ali Farhadi$^\psi$, Kiana Ehsani$^\dagger$, Aniruddha Kembhavi$^{\dagger\psi}$}\\
$^\dagger$PRIOR @ Allen Institute for AI, $^\psi$University of Washington, Seattle \\
\href{https://allenai.org/project/phone2proc}{https://allenai.org/project/phone2proc}
}

\twocolumn[{
\renewcommand\twocolumn[1][]{#1}
\maketitle
\vspace*{-0.5cm}
\centering
\includegraphics[width=0.95\linewidth]{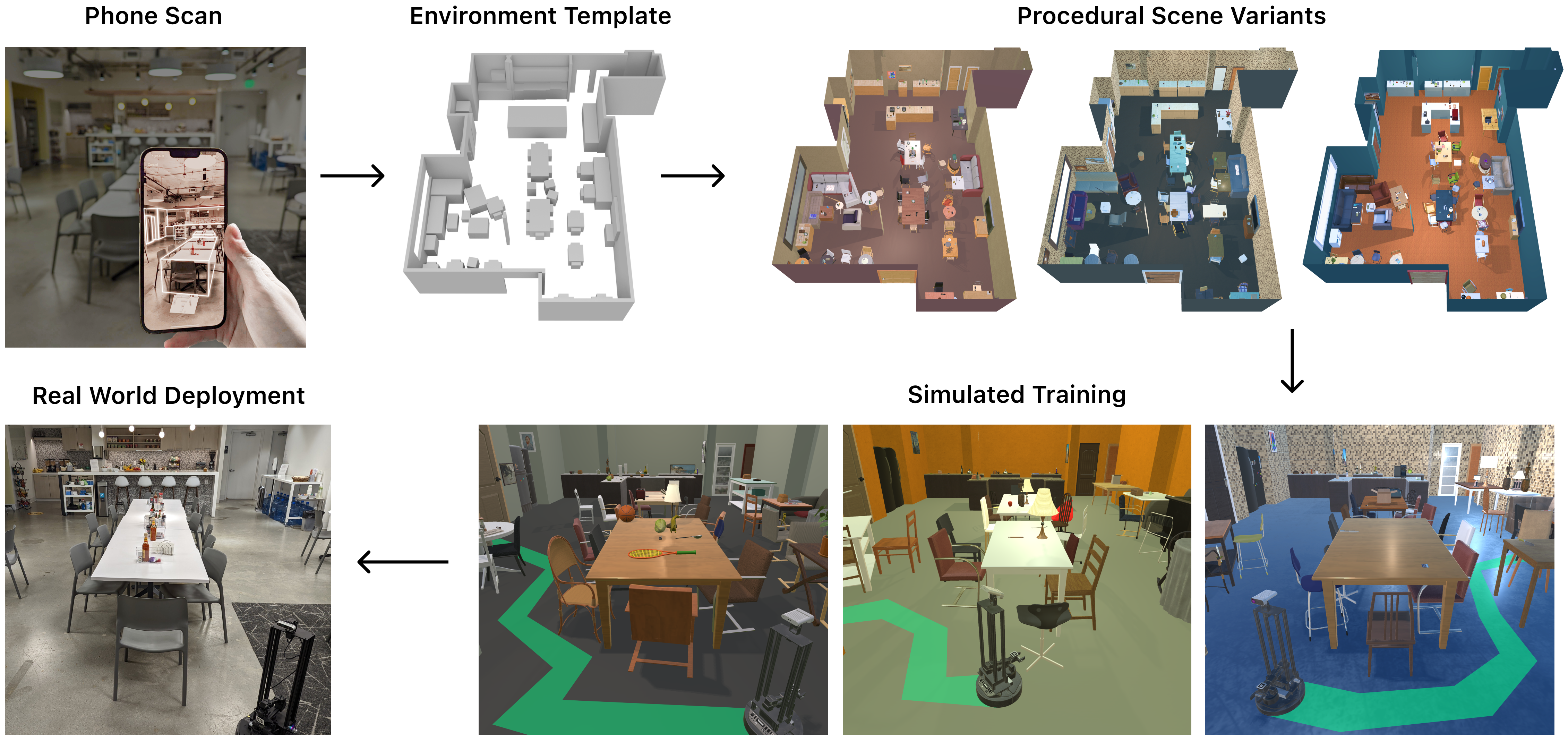}
\vspace{-.2cm}
\captionof{figure}{Successfully deploying agents trained in simulation to the real world has generally proved fraught - we present \methodName, a simple approach that uses a cellphone to scan an environment and procedurally generate targeted training scene variations of that location, whose usage results in successful and robust agents in the real environment.}
\label{fig:teaser}
\vspace*{0.3cm}
}]

\maketitle

\blfootnote{$^*$ Equal contribution.}

\begin{abstract}
   \vspace{-0.5em}

Training embodied agents in simulation has become mainstream for the embodied AI community. However, these agents often struggle when deployed in the physical world due to their inability to generalize to real-world environments. In this paper, we present Phone2Proc, a method that uses a 10-minute phone scan and conditional procedural generation to create a distribution of training scenes that are semantically similar to the target environment.
The generated scenes are conditioned on the wall layout and arrangement of large objects from the scan, while also sampling lighting, clutter, surface textures, and instances of smaller objects with randomized placement and materials.
Leveraging just a simple RGB camera, training with Phone2Proc shows massive improvements from 34.7\% to 70.7\% success rate in sim-to-real ObjectNav performance across a test suite of over 200 trials in diverse real-world environments, including homes, offices, and RoboTHOR.
Furthermore, Phone2Proc's diverse distribution of generated scenes makes agents remarkably robust to changes in the real world, such as human movement, object rearrangement, lighting changes, or clutter.

\vspace{-1em}
\end{abstract}

\section{Introduction}
\label{sec:intro}
The embodied AI research community has increasingly relied on visual simulators~\cite{Kolve2017AI2THORAI,Savva2019HabitatAP,Xia2018GibsonER} to train embodied agents, with the expectation that the resulting policies can be transferred onto robots in the physical world.
While agents trained within simulated environments have shown increased capabilities, progress in successfully deploying these policies onto physical robots has been limited.

Robots trained in simulation must overcome daunting challenges if they are to work effectively in a real space such as our home. First, they must overcome the generalization gap between the limited set of simulated environments they are trained on and the test scene of interest. In practice, policies trained to perform complex visual tasks with reinforcement learning struggle to perform well in novel scenes with novel layouts and object instances. Second, they must work in realistic environments where we live and work, which are often full of clutter, with objects that keep being moved around, with people in and out of the scene and with lighting changes. In short, we expect our agents to learn from a small set of training data points and generalize not just to a single test data point, but to a distribution of test data that is often semantically distant from the training data. Today's methods are a ways away from delivering such performant, robust, and resilient robots~\cite{deitke2020robothor,Chattopadhyay2021RobustNavTB}.

In this work, we present \methodName, which represents a significant advancement towards the goal of creating performant, robust, and resilient robots. Instead of training policies in simulated environments that may be semantically distant from the target physical scene, \methodName efficiently generates a distribution of training environments that are semantically similar to the target environment. This significantly reduces the generalization gap between the training and target distributions, resulting in more capable robots.

\methodName utilizes a freely available mobile application to quickly scan a target environment and create a template of the surroundings, including the scene layout and 3D placements of large furniture. This template is then used to conditionally generate a fully interactive simulated world using ProcTHOR~\cite{deitke2022procthor}, closely mirroring the real-world space. Importantly, this single simulated environment is then transformed into a distribution of simulated worlds by randomizing objects, their placements, materials, textures, scene lighting, and clutter. This allows for the creation of arbitrary large training datasets that are semantically similar to the desired real-world scene.

We produce policies for object goal navigation using \methodName and deploy them onto a LoCoBot robot in the physical world. We conduct extensive evaluations with 234 episodes in five diverse physical environments: a 3-room and 6-room apartment, a test scene from RoboTHOR-real, a conference room, and a cafeteria. This represents one of the largest and most diverse studies of sim-to-real indoor navigation agents to date. Across all environments, \methodName significantly outperforms the state-of-the-art embodied AI model built with ProcTHOR, with an average improvement in success rate from 34.7\% to 70.7\%. Our robot is able to explore the scene efficiently and effectively navigate to objects of interest, even in the presence of clutter, lighting changes, shifts in furniture, and human movement. These strong navigation results are achieved using an \textbf{RGB-only camera}, \textbf{no depth} sensors, \textbf{no localization} sensors, and \textbf{no explicit mapping} components.

In summary, we present: (1) \methodName, a simple and highly effective method for reducing the generalization gap between datasets of simulated environments and a target environment in the real world, (2) large-scale real-world robotics experiments with 234 trials showing significant improvements for \methodName compared to state-of-the-art models, and (3) experiments demonstrating the robustness of \methodName in the face of variations such as changes in lighting, clutter, and human presence.

\section{Related Works}
\label{sec:related}
\begin{figure*}[t!]
    \centering
    \includegraphics[width=\textwidth]{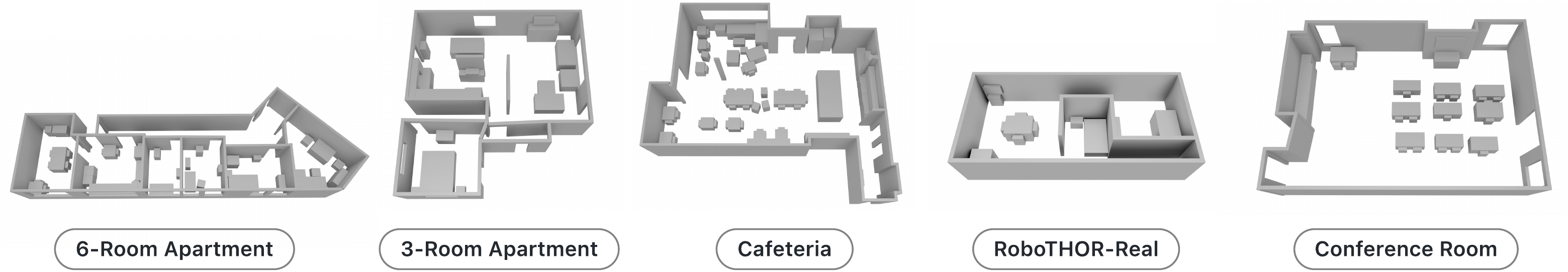}
    \caption{\textbf{Examples of environment templates for our five target test environments.} These are produced by an iPhone scanning each environment using our iOS app that leverages Apple's RoomPlan API. These environment templates contain the room layouts and some 3D locations for large furniture objects. They do not contain small objects, textures, lighting, etc. }
    \label{fig:scan-sample}
    \vspace{-1em}
\end{figure*}

\mypara{Navigation in Simulated Environments.} 
Visual navigation~\cite{Batra2020ObjectNavRO,Anderson2018OnEO} is a popular task in the embodied AI community with benchmarks in several simulators~\cite{Kolve2017AI2THORAI,Xia2018GibsonER,Savva2019HabitatAP,Ramakrishnan2021HabitatMatterport3D}. An effective approach is to use semantic mapping to explore environments efficiently~\cite{Chaplot2020ObjectGN,chaplot2020learning,Chaplot2020NeuralTS,Georgakis2022LearningTM}. Kumar \etal\cite{Kumar2022GoalConditionedEF} adapts mapping methods to condition the policy on the target. These works utilize agent pose and depth sensors to build their maps and localize the agent. In contrast, our method only relies on RGB information without any additional sensor. Other methods for navigation use scene priors~\cite{Yang2019VisualSN}, meta-learning~\cite{Wortsman2019LearningTL}, paired grid world environments~\cite{Jain2021GridToPix}, scene memory transformers~\cite{Fang2019SceneMT}, passive videos of roaming in a scene as a training cue~\cite{Hahn2021NoRN} and expert human trajectories for imitation learning~\cite{Ramrakhya2022HabitatWebLE}.

The community has also made progress in training embodied agents for navigation exclusively using RGB observations and barebones neural architectures. These include using frozen ImageNet trained visual encoders~\cite{Zhu2017TargetdrivenVN}, learning visual encoders from scratch~\cite{deitke2020robothor}, and using CLIP~\cite{Li2021ACM} based encoders~\cite{Khandelwal2022SimpleBE}. Gadre \etal\cite{Gadre2022CLIPOW} and \cite{Majumdar2022ZSONZO} use an off-the-shelf exploration method and clip-based object localization to accomplish 0-shot object navigation. Deitke \etal\cite{deitke2022procthor} show the benefits of procedural generation  for navigation and manipulation. 

While the above works show promising results in simulation, most are not deployed and tested on real robots. We provide comparisons to the current state-of-the-art method, ProcTHOR~\cite{deitke2022procthor}, via large-scale real-world evaluations.

\mypara{Sim-to-Real Transfer.} While most models are evaluated only in simulation, for practical applications, policies learned in simulation must function in real life. Often, policies trained only in simulation can prove brittle or nonfunctional in transfer \cite{jakobi1995noise}. Chatopadhyay \etal\cite{Chattopadhyay2021RobustNavTB} find that standard embodied agents significantly underperform (or fail) in the presence of realistic noise in the system. 
Truong \etal\cite{Truong2022RethinkingSL} compare the correlation of the performance of 4-legged robots navigating in simulation against the real world. They find that adding fidelity to the simulation does not help with the performance in the real world. 

An alternate approach to higher fidelity is to add randomization to sensing or dynamics in simulation. This does help \cite{tobin2017domain,sadeghi2016cad2rl}, but too much randomization can degrade training efficacy \cite{matas2018sim}, and hand-tuning appropriate randomization requires expert knowledge and does not scale. Some address this pitfall by leveraging real-world rollouts or inputs at train time to tune simulation randomization \cite{du2021auto,chebotar2019closing}. However, these works are randomizing a subset of a well-parameterized dynamical system for a narrow task (swing-peg-in-hole or cabinet opening), as opposed to a more open-ended task or randomizing the entire visual appearance and object instances of the environment.

Recent works have deployed and measured policies on real robots~\cite{Kadian2020Sim2RealPD,Bigazzi2021OutOT,Sadek2022AnIE,Dugas2022NavDreamsTC,Truong2021BiDirectionalDA} for the task of point goal navigation. 
In contrast to most, we use no mapping, explicit localization, or depth, as well as targeting a more complex task. We also test more extensively and in a wider variety of environments. Anderson \etal\cite{Anderson2020SimtoRealTF} study the sim-to-real gap for vision and language navigation and discover a crucial need for an oracle occupancy map and navigation graph. Our method does not require a manually annotated map and is robust to moving obstacles without the necessity for an additional dataset.

\mypara{Real-to-Sim Transfer.} Transferring observations from the real world to simulation can open up further capacities for training agents. This has been studied in the domain of object manipulation~\cite{Lim2022Real2Sim2RealSL,Wang2022ARM}. 
\cite{Ehsani2020UseTF} replicate an observed manipulation trajectory by predicting contact points and the forces.
\cite{Jiang2022DittoBD,Nie2022StructureFA} generate a 3D mesh of an object with articulation from observed interactions.
\cite{Sundaresan2022DiffCloudRF,Antonova2022ABT} infer simulation parameters for deformable object manipulation. \cite{Yang2017LearningBasedCM,Wang2011DatadrivenEM} learn cloth material recovery from videos. Our focus is on conditioning our procedural generation on the real scene rather than perfectly replicating the observation. \cite{Prakash2021SelfSupervisedRS} use a self-supervised technique to utilize unlabeled images for acquiring data for training scene graph models. We focus on generating 3D interactable environments for training embodied agents.

\mypara{Navigation in Robotics.}
The robotics community has made progress in navigating robots with different embodiments in a diverse set of environments. Gupta \etal~\cite{Gupta2017CognitiveMA} combine a differentiable planner module with mapping to train a visual navigation agent end-to-end. In contrast to our work, their method assumes perfect odometry. Different methods have been used to build agents that follow demonstrated paths and trajectories or navigate \cite{Hirose2019DeepVM,Kumar2018VisualMF,Meng2021LearningCB,Meng2020ScalingLC,Xie2020SnapNavLM,Savinov2018SemiparametricTM}. Shah \etal~\cite{Shah2022GNMAG,Shah2021RECONRE} build models for open world navigation.
The main focus of these works is on the low-level control systems of the robots, whereas our focus is on building end-to-end models for embodied visual reasoning.

\section{Approach}
\label{sec:approach}

We now present \methodName which generates a distribution of training environments that closely match the real world physical space we are interested in. We begin with a phone scan of a target scene (Sec~\ref{sec:phone}), then condition on this scan to procedurally generate variations of the scene for training agents (Sec~\ref{sec:2proc}), and finally transfer onto a LoCobot robot that navigates in the physical world (Sec~\ref{sec:2real}).

\subsection{Scanning}
\label{sec:phone}

\methodName\ is designed to optimize a robot's performance within a desired real world environment. The first step in this process is to scan the target environment. This is accomplished using an iOS app that we built and will release using Apple's freely available RoomPlan API~\cite{roomplan}. Scanning a large apartment with several rooms only takes a few minutes, can be done using an iPhone or iPad and it outputs the environment template as a USDZ file.

The RoomPlan API provides us with a high-level bounding box template of the environment, which contains the room layouts and 3D placement of large objects that are visible to the camera. While scanning an environment, the app provides detailed real-time feedback about the construction of the scene to help the user capture a more accurate scan.
 
The resulting environment template includes the 3D locations and poses of walls, large objects, windows, and doors. Each object in the scan is assigned to one of 16 object types, including storage, sofa, table, chair, bed, refrigerator, oven, stove, dishwasher, washer or dryer, fireplace, sink, bathtub, toilet, stairs, and TV. Smaller objects, such as those typically on surfaces, are ignored. The metadata produced for each object includes the size, position, and rotation of its 3D bounding box, along with the forward-facing orientation. %
Doors and windows are provided as cutouts in the walls. Figure~\ref{fig:scan-sample} presents examples of scanned environments showcasing the diversity of the layouts in our test set.

\subsection{Environment-Conditioned Procedural\\ Generation}
\label{sec:2proc}

Procedural generation of simulation environments allows for a vast diversity of scenes for agents to train on. Deitke \etal~\cite{deitke2022procthor} begin with a high-level room specification (\eg 2 bedroom house with a kitchen and living area) and create an environment that matches it. In contrast, our approach uses a scan of the target real-world environment to condition the generation and create variations of that scene.

\methodName\ parses the USDZ environment template produced by the iOS app and extracts wall, door and window positions and 3D large object bounding boxes. It then leverages ProcTHOR to generate a fully rendered scene in Unity and finally populates this scene using ProcTHOR's asset database of 1,633 assets across 108 object types. The generation process is very fast and can generate 1000 procedural scene variants in around an hour with an 8 Quadro RTX 8000 GPU machine.

While Deitke \etal~\cite{deitke2022procthor} generates and populates scenes starting from a high-level room specification (e.g. a 2 bedroom house with a kitchen and living area), our approach generates scene variations that are conditioned on a scan of the target real-world environment. This process involves (a) parsing the environment template, (b) generating the scene layout, (c) sampling objects from the asset library to match scanned semantic categories, (d) accounting for object collisions in Unity, (e) populating the scene with small objects not captured by the scan, and (f) assigning materials and lighting elements. We now provide further details on each of these steps.

\begin{figure}[tp]
    \centering
    \includegraphics[width=20pc]{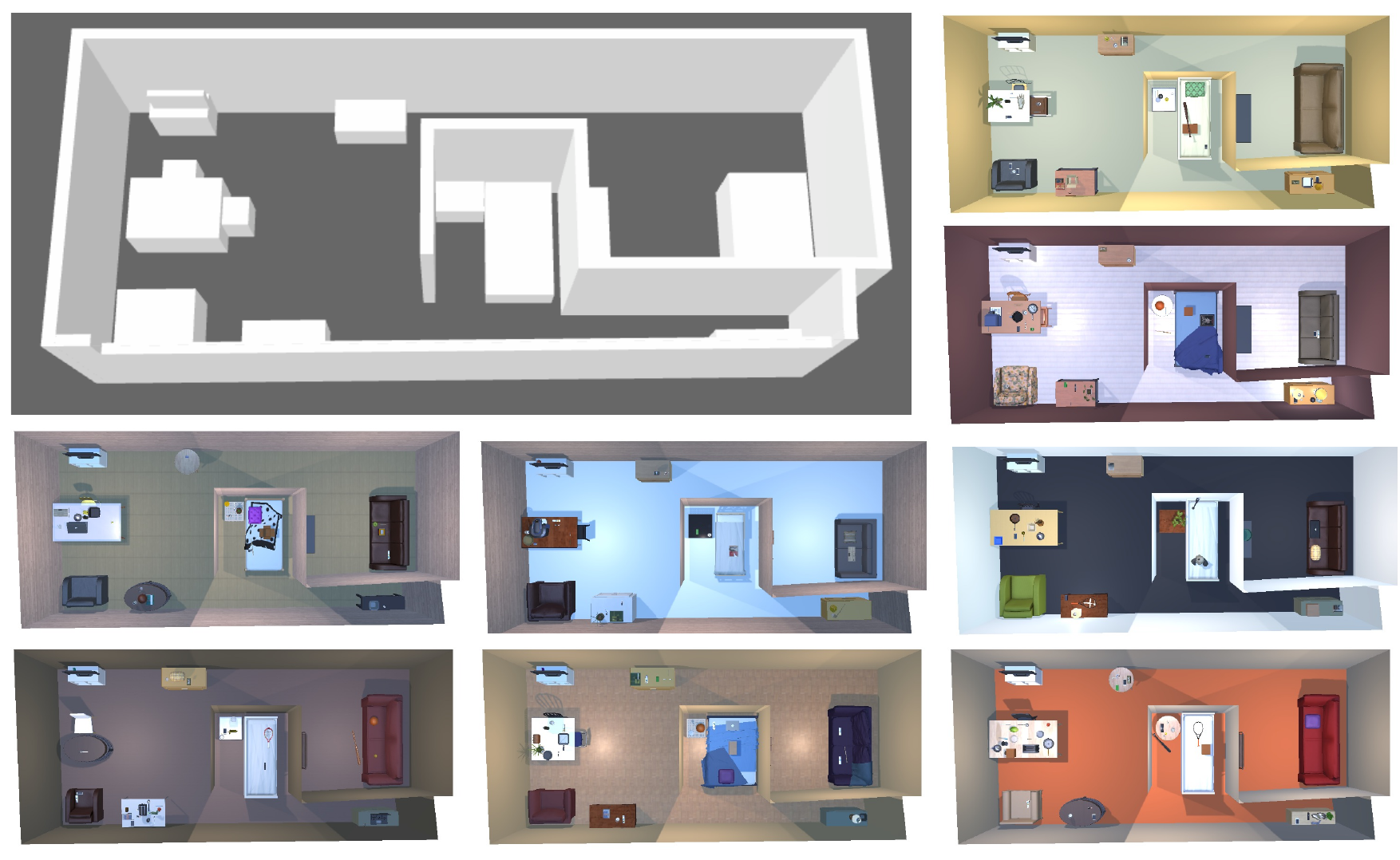}
    \caption{\textbf{Examples of Procedurally Generated Houses.} The procedural generation of the houses is conditioned on the target environment scanned using a phone. We are able to sample a rich and diverse set of scenes from this distribution with varying lighting, textures, objects and placements.}
    \label{fig:phone2proc-samples}
\end{figure}

\noindent \textbf{Layout.} The environment specification file contains the placement of walls in each room. 
Unlike walls generated in ProcTHOR-10K~\cite{deitke2020robothor}, which are only aligned to orthogonal axes, \methodName allows for a more diverse wall generation that can accommodate any scanned layout. Each wall's specification comes from its 3D bounding box, width, height and a constant depth (\eg 16cm for all walls) -- which are used to produce a wall asset within the simulated environment. Placing walls produces the external boundary of the environment as well as its internal layout.

\noindent \textbf{Rooms.} We partition the space located within the external boundary walls into distinct rooms. A room is formed if the walls formed from the top-down 2D plane fully enclose a polygon. This is followed by floor and ceiling generation.

\noindent \textbf{Windows and doors.} The environment template specifies if each wall has cutouts for windows and doors. Our USDZ parser extracts the size and position of the holes along the wall. If the hole includes a cutout at the bottom of the floor, we place a door there; otherwise, we place a window. Here, we uniformly sample a door or window asset from the asset database and scale it appropriately.

For doors between connecting rooms, the sampled door may either include just a frame or both a frame with an openable door and a degree of openness sampled uniformly between 0.8 and 1.0. The room that the door opens into is randomly sampled. If the door is connected to the outside of the environment, we sample a door frame with an openable door component and fully close the door (to prevent agents from getting out of the scene).

\noindent \textbf{Semantic objects.} For each object in the template, we wish to sample an appropriate asset matching its semantic category. For each semantically similar ProcTHOR object candidate, we compute its 3D bounding box IoU with the object it may represent in the environment template and reject candidates with an IoU less than 75\%. We then uniformly sample from the rest. The sampled asset's position on the floor and its forward-facing direction come from the environment template's corresponding object. We compute the vertical position of the object based on if it is on a surface (\eg a couch on the floor or a television on top of a table) or attached to a wall (\eg a wall television).

This procedure to find a matching asset can sometimes lead to large variations. For example, a table object may match a ProcTHOR coffee table, side table, or dining table and a TV may match a flat-screen TV or a vintage box television. Randomly sampling different asset instances in the library of a particular semantic object type makes agents more robust as they must learn to generalize to many visually distinct instances for each type.%

\noindent \textbf{Object collisions.} We check to make sure that none of the placed objects collide with one another or the walls in the scene to avoid unrealistic configurations. %

\noindent \textbf{Small objects.} After placing the large objects that match the scan, we generate smaller objects to be placed on top of them. Here, for instance, we might populate a bed with pillows or place fruits and plates on the counter. Unlike what happens in a 3D reconstruction, where all the objects are static, we are able to randomize the placement of small and target objects to produce many scene variations and prevent overfitting (See experiments in Sec~\ref{sec:experiment}).

\noindent \textbf{Lighting.} Scene lighting is randomized such that each room is guaranteed one light, and then additional lights are uniformly sampled throughout the scene scaled to the number of rooms. Each light is then randomized in its intensity, RGB values, shadow bias, and strength.

\noindent \textbf{Materials.} We randomize materials following ProcTHOR's material randomization by sampling from sets of structure materials (\ie wall, floor, and ceiling) and object materials.

\noindent \textbf{Clutter.} After all the semantic objects from the scan have been sampled and smaller objects have been sampled to be placed on top of them, we also sample additional clutter objects (such as boxes, dumbbells, pillows, and plants) that are placed on the floor of each room -- similar to how a real house may have objects like kids toys thrown around. These objects prevent overfitting to particular paths in the environment and helps teach agents to avoid obstacles.

\subsection{Transfer to Real World}
\label{sec:2real}

We first detail our model architecture and training regime then discuss our robot and physical scenes.

\noindent \textbf{Model and Training Details.}
\label{sec:model}
Our goals are to: a) design models that do not depend on unrealistic sensory data in real indoor environments like agent or target localization, b) use only RGB observations since real world depth cameras offer few choices, generally come with small FOV and are fairly noisy, and c) create agents that are robust to clutter and changes in the environment.

\methodName\ provides a distribution of simulated worlds that are sampled to produce a large training set. These scenes differ in the placement of small objects, materials, lighting, clutter, etc. This allows us to train policies that do not overfit to a single scene configuration, but instead generalize to realistic scene variations.

\begin{figure}[tp]
    \centering
    \includegraphics[width=\columnwidth]{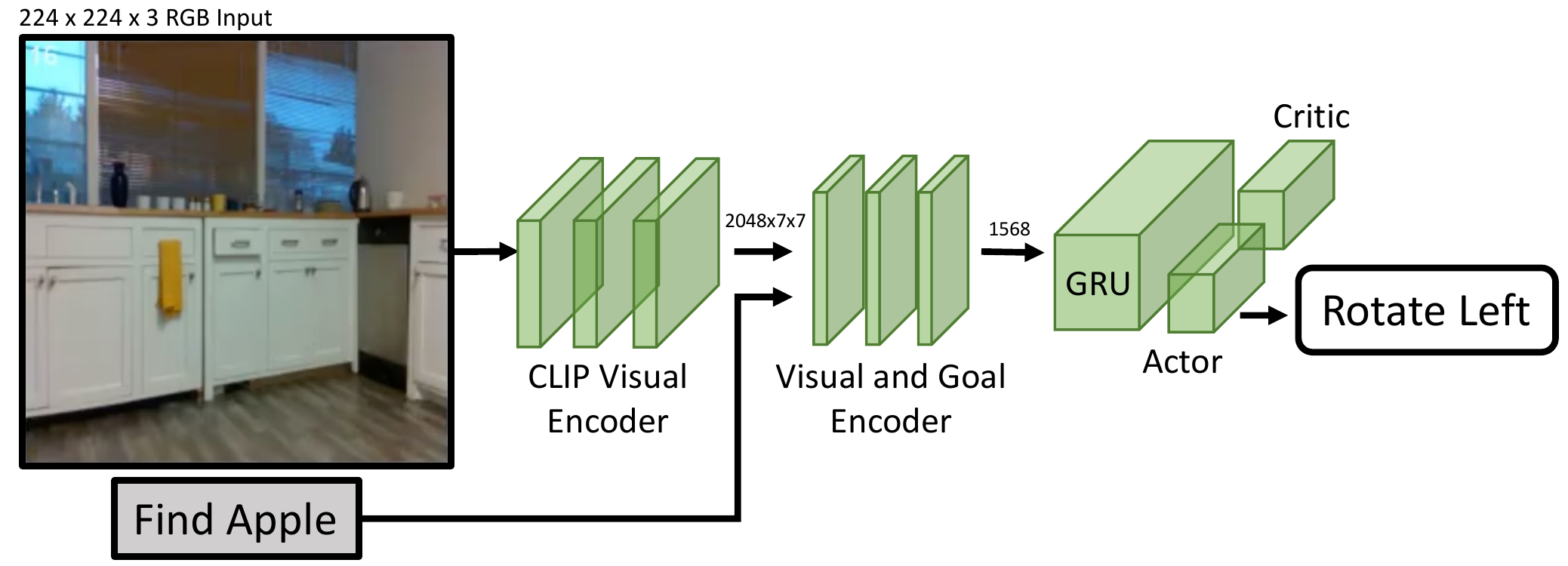}
    \caption{Our architecture is a simple GRU operating on the CLIP encoding of solely RGB input.}
    \label{fig:model}
\end{figure}

In terms of the model design, we adopt a simple architecture introduced in \cite{Khandelwal2022SimpleBE} and also used in \cite{deitke2022procthor}. The model uses a CLIP encoder to embed the visual observation (ego-centric RGB frame) followed by a GRU to capture temporal information (Fig \ref{fig:model}).  We pre-train our model on the ProcTHOR-10k dataset using the same training regime presented in \cite{deitke2022procthor} and then finetune on the Phone2Proc environments for the task of object navigation on 16 object categories. We use AllenAct~\cite{AllenAct} to train our models. More details on the training pipeline are provided in the appendix.

In principle, it is fairly straightforward to adopt a more complex model architecture, but we found this simple design to be highly performant not just in simulation but also in our real world experiments. Similarly, it is also easy to train agents for other tasks, including one involving object manipulation using an arm, since \methodName\ produces scenes that are fully interactive with support for all agents in AI2-THOR~\cite{Kolve2017AI2THORAI} including the arm-based agent~\cite{Ehsani2021ManipulaTHORAF}.

During fine-tuning, we lower the learning rate to $0.00003$ to avoid catastrophic forgetting of the skills learned in pre-training. We add a failed action penalty ($0.05$) in the reward shaping to encourage the agent to avoid hitting the obstacles in the environment. This is especially important as we deploy these models in the real world and would like to avoid damage to the environment or the robot. Instead of hand-tuning the camera parameters to match perfectly with the real world, the FOV of the camera in simulation is randomly sampled from a distribution approximating the real world.

\noindent \textbf{Real-World Experiments.}
\label{sec:2real-realworld}
Models trained on procedurally-generated variants of the scene scans are then directly evaluated in real environments. We use 5 environments: a 3-room apartment, a large 6-room apartment, a real world test scene from RoboTHOR \cite{deitke2020robothor}, a large re-configurable office conference room, and a cafeteria. Models are evaluated against 5 different target objects from 3 different starting locations in the environment. No training or calibration is performed in the real world.

No particular effort was made to arrange for ease of robotic experimentation. The goal was to use real environments in their most natural setting. The lighting, object instances, textures, and window views are not recognized by the \methodName\ scan and are thus unseen by the agent at training time. As there are many objects in the real-world scenes that are not present in ProcTHOR's asset library (\eg whiteboards, bicycle), there are several object categories in each environment that are novel to the agent. No additional information is used in the preparation/scanning step besides the output of the RoomPlan API. 

Experiments are run on LoCoBot \cite{gupta2018robot}, a low-cost, platform about 60cm tall using the PyRobot API \cite{murali2019pyrobot}. The agent's discrete action space is look up/down, turn right/left (each 30$^o$), move ahead 25cm, and a ``done'' action to indicate reaching the target. Actions are sampled using PyTorch's categorical distribution. FPS in real is $\approx$0.25, and for practical reasons, physical trajectories were limited to 250 or 500 steps depending on the size of the environment. %

\section{Experiments}
\label{sec:experiment}
We provide extensive real-world evaluations of \methodName. In Sec.~\ref{sec:how-well} we compare \methodName to \procthor in 5 diverse real environments. In Sec.~\ref{sec:upperbound} we show that \methodName performs as well as a privileged upper bound setting that utilizes a simulated counterpart of the real-world environment, painstakingly modeled by a digital artist. Sec.~\ref{sec:robustness} illustrates the robustness of \methodName to various realistic changes in the environment, showing how \methodName hugely improves over using static reconstructions. Finally, we statistically analyze the significance of our results (Sec~\ref{sec:stat-analyze}).

\noindent \textbf{Scale of real-world evaluations.} In aggregate we conduct 234 episodic evaluations in 5 diverse real-world environments. Our environments are large and challenging and each episode takes between 5 and 20 minutes to run. This represents one of the largest and most diverse real-world evaluation studies of sim-to-real indoor navigation agents. We put this number in the context of related works that provide 20 trials (1 scene)~\cite{Chaplot2020ObjectGN}, 1 qualitative example~\cite{chaplot2020learning}, 36 trials (1 scene)~\cite{deitke2020robothor} and 9 episodes (1 scene)~\cite{Partsey2022IsMN}. A recent study for PointNav for studying Sim-vs-Real correlation~\cite{Kadian2020Sim2RealPD} conducts 405 real trials but only uses a single laboratory setting.

\noindent \textbf{Training models and baselines.}
All models use the same architecture and begin from a checkpoint trained on ProcTHOR-10k train set for the task of object navigation. This checkpoint is state of the art on 6 benchmark Embodied AI tasks \cite{deitke2022procthor}. This checkpoint is then fine-tuned for 5M steps on ProcTHOR-10K train with modifications detailed in section \ref{sec:model}, and is henceforth referred to as \procthor or ``baseline''. The \methodName models are environment-specific and are fine-tuned on 1K procedurally generated variants of scans of the relevant environment. Results for these are presented in Figure \ref{fig:four_env_results}.

\subsection{How Well Does Phone2Proc Work?}
\label{sec:how-well}

We evaluate \methodName in 5 diverse real-world environments: a 3-room apartment, a 6-room apartment, 1 RoboTHOR-Real apartment, a conference room, and a cafeteria. In each space, our model and the baseline are each evaluated for 15 trials (5 object categories with 3 agent initial locations per category). The 5 categories (apple, bed, sofa, television, and vase) were chosen to showcase both fixed objects whose locations can be learned (\eg television) and small objects that must be searched for (\eg apple). Where necessary (\eg conference rooms do not usually contain beds), the bed and sofa are substituted for environment-appropriate objects such as chairs or garbage cans. Starting locations are geographically distributed and we avoid ones that would achieve trivial success.

\begin{figure}[tp]
    \centering
    \includegraphics[width=20pc]{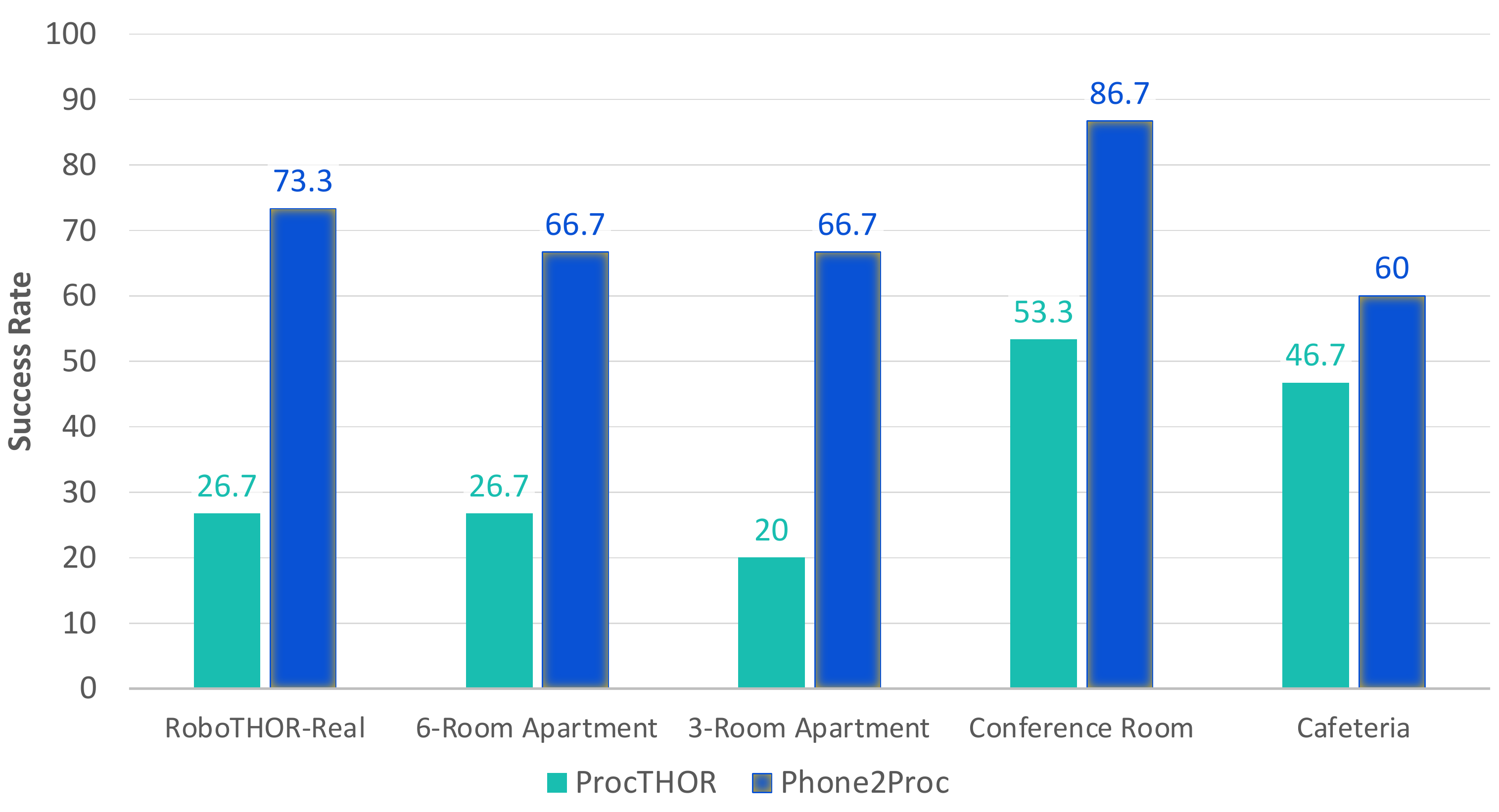}
    \caption{Results for \methodName vs ProcTHOR baseline in a variety of real environments. Each number represents fifteen trajectories - five objects from three starting locations.}
    \label{fig:four_env_results}
    \vspace{-1em}
\end{figure}

Fig.~\ref{fig:four_env_results} shows that in every real-world scene, \methodName performs remarkably well and significantly outperforms the \procthor baseline. In aggregate, \methodName achieves a Success Rate of 70.68\% compared to 34.68\% for \procthor. Overall, we find that the bigger environments with multiple rooms (Robothor, 3 room apartment and 6 room apartment) are quite challenging for the baseline. \methodName on the other hand, performs very effectively in these scenes, that require it to perform long range exploration.

\begin{table}[h]
\footnotesize
\setlength{\tabcolsep}{2pt}
	\centering
\hfill
	\begin{tabular}{l|cc}
	\toprule
    Model    & Success Rate & Episode Length \tabularnewline%
\midrule 
Habitat~\cite{Ramakrishnan2021HabitatMatterport3D}& 33.3 & 204.8  \tabularnewline%

\procthor~\cite{deitke2022procthor}& 33.3 & 92.5  \tabularnewline%
\midrule
\methodName (ours) & \textbf{77.8}  &\textbf{ 82.6}  \tabularnewline%

\bottomrule

	\end{tabular}
\hfill 
	\vspace{0cm}
	\caption{Results of 9 trajectories evaluated in RoboTHOR-Real.}
	\label{table:robothor_results}
\end{table}

Table~\ref{table:robothor_results} compares \methodName with the same model architecture trained on Habitat~\cite{Ramakrishnan2021HabitatMatterport3D} (implementation details on baseline training in the appendix). These results are presented on RoboTHOR-Real for 9 (instead of 15) episodes since Habitat only covers 3 of the 5 objects in our target set (bed, sofa, and television).

\begin{figure*}[tp]
    \centering
    \includegraphics[width=40pc]{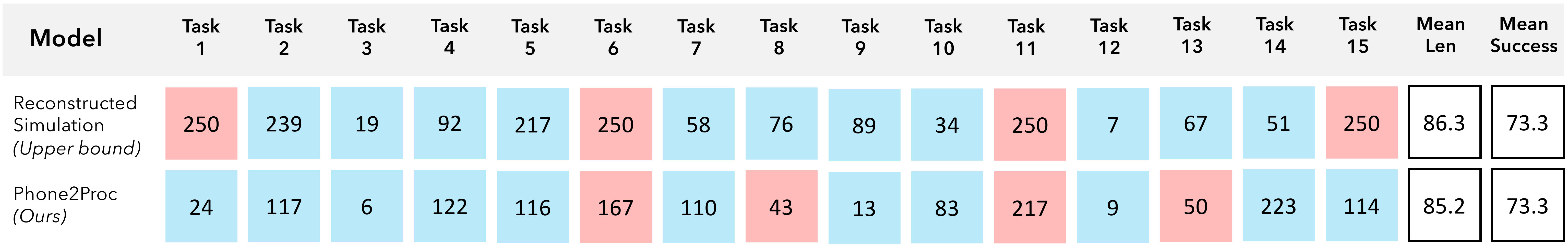}
    \caption{\textbf{Comparison with the reconstructed simulation.} We compare \methodName with the privileged reconstruction baseline among 15 different tasks (each task represents a different pair of agent's initial location and target object). Each square shows the episode length of the corresponding model for each task. The red squares represent failed episodes and the blue ones indicate successful ones. Despite the baseline's privileges, \methodName achieves a similar success rate and episode lengths.}
    \label{fig:onesim}
    \vspace{-1em}
\end{figure*}

\subsection{How Does Phone2Proc Compare To A\\ Privileged Upper Bound?}
\label{sec:upperbound}
RoboTHOR test scenes come with a carefully and manually reconstructed simulation counterpart. This allows us to train a privileged model on this perfect replica. Producing this replica for a real scene took 5 days and is intractable practically but represents a theoretical upper bound, in terms of quality and correctness, for static 3D reconstruction methods that may employ sensors such as LiDAR. The model, referred to as \emph{\onesimlong} or \onesim, is only finetuned in this 1 scene. In simulation, \onesim achieves 100\% success since it overfits that scene easily.

We evaluate \methodName and \onesim for 15 episodes in the RoboTHOR-real apartment (Fig. \ref{fig:onesim}). \methodName is able to match the performance of this privileged baseline both in terms of Success and Episode length, which shows the effectiveness of our proposed approach to scan the target environment and train within its variations. In Sec \ref{sec:robustness}, we show the pitfalls of this privileged model.

\begin{figure*}[h!]
    \centering
    \includegraphics[width=40pc]{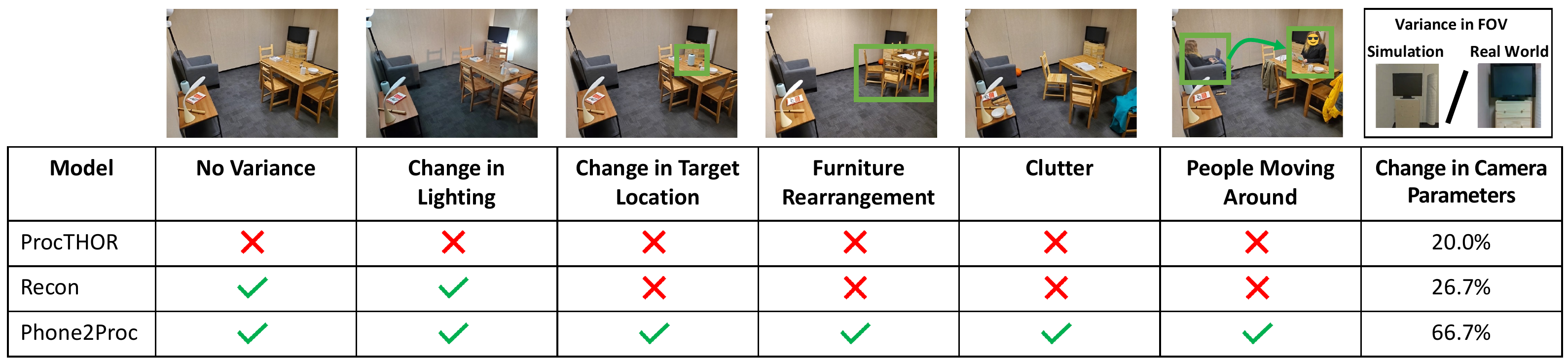}
    \caption{Illustration of the scene disturbances used to comparatively evaluate models, along with their performances over each episode. The third column showcases the performance of models navigating to a vase when the object's location is changed. For the last column, a full set of 15 trajectories is evaluated for each model. For other disturbances, we evaluated models for the target of Television. %
    }
    \label{fig:fragility}
\end{figure*}

\subsection{How Robust is Phone2Proc to Chaos?}
\label{sec:robustness}

In reality, our homes and offices aren't static and picture-perfect. Objects move around, furniture gets shifted, kids leave their toys on the floor, people keep moving around in the scene, lighting keeps changing throughout the day, and more! We evaluate the baseline model \procthor, the privileged model \onesim and our model \methodName in these settings (Fig.~\ref{fig:fragility}). 

First, \procthor\ does poorly in all settings, unsurprising given that it also fails on the episode with no variation. \onesim performs well with no variations (consistent with Fig. \ref{fig:onesim}). However, it performs very poorly when variations are introduced in the scene. When objects are moved around by just 1.5m, \onesim fails, as it has memorized the location of every target object. Clutter and chair position adjustment confuses it, and the agent is simply unable to move around the scene and explore effectively. Moving the dining room furniture closer to the wall, as one might in a  real house, produces interesting behavior. \onesim calls the \emph{Done} action for the television target when it sees the lamp. This is because in the original scan, the lamp was next to the television, and this is what the model likely memorized. \onesim also fails when people move around during an episode. In stark contrast, \methodName is robust to every variation we tested, showing that procedurally generating variations of the scan helps train robust agents.

Finally, we tested all three models for robustness towards a change in camera parameters between simulation and the real robot. A full 15 trajectories were evaluated for each model trained with a wide vertical FOV and evaluated with a narrow one. \methodName is robust to this change, while \onesim's performance drops drastically.

\subsection{Statistical Analysis}
\label{sec:stat-analyze}

As described above, we have jointly evaluated \textsc{Phone2Proc} and \textsc{ProcTHOR} models across 3 starting positions in 5 real environments with 5 target objects per environment (chosen from 7 unique types). Together this amounts to (2 model types) ${\times}$ (3 positions) ${\times}$ (5 environments) ${\times}$ (5 targets) $= 150$ datapoints. In order to validate the statistical significance of our results, we follow a similar analysis as in~\cite{Weihs2021HideAndSeek} and model agent success using a logistic regression model in \texttt{R}~\cite{R}. In particular, here we model all exogenous variables as fixed effects and, as starting positions are inherently nested within environments, we include all environment and starting position interactions. When fitting this model, we obtain the coefficient estimates\\[-2.5em]
\begin{center}
\resizebox{1\linewidth}{!}{%
\begin{tabular}{l|cccccccccc}
        & (Intercept) & Phone2Proc & Bed & Chair & GarbageCan & Sofa & TV & Vase \\ \hline 
Coef.           & 0.17 & 0.33        & 0.00 & 0.41 & -0.12 & 0.2 & -0.03 & 0.13  \\ 
$p$-value       & 0.31 & ${<}$0.0001 & 0.97 & 0.02 & 0.59  & 0.13 & 0.78 & 0.27 \\
\end{tabular}
}
\end{center}
~\\[-0.5em]
where, for space, we have excluded coefficients corresponding to environments and locations, as none of these coefficients were statistically significant at a 0.05 level. As the above shows, the coefficient of interest (\methodName) is statistically significant even at a $0.0001$ level. Interpreting these results, we see that when holding other factors constant, the use of a \methodName model is associated with $\exp(0.33)= 1.39$ times greater odds of success (95\% confidence interval: $[1.2, 1.62]$) than when using a \textsc{ProcTHOR} model. Of all object categories, only the coefficient associated with chair was found to be statistically significant at a 0.05 level suggesting that chairs were associated with higher levels of success (\ie may be generally easier to find). Altogether, we find strong statistical evidence suggesting that, across tested environments and object categories, \methodName was associated with higher success rates and that the estimated effect size was substantial.

\begin{figure}[ht!]
    \centering
    \includegraphics[width=20pc]{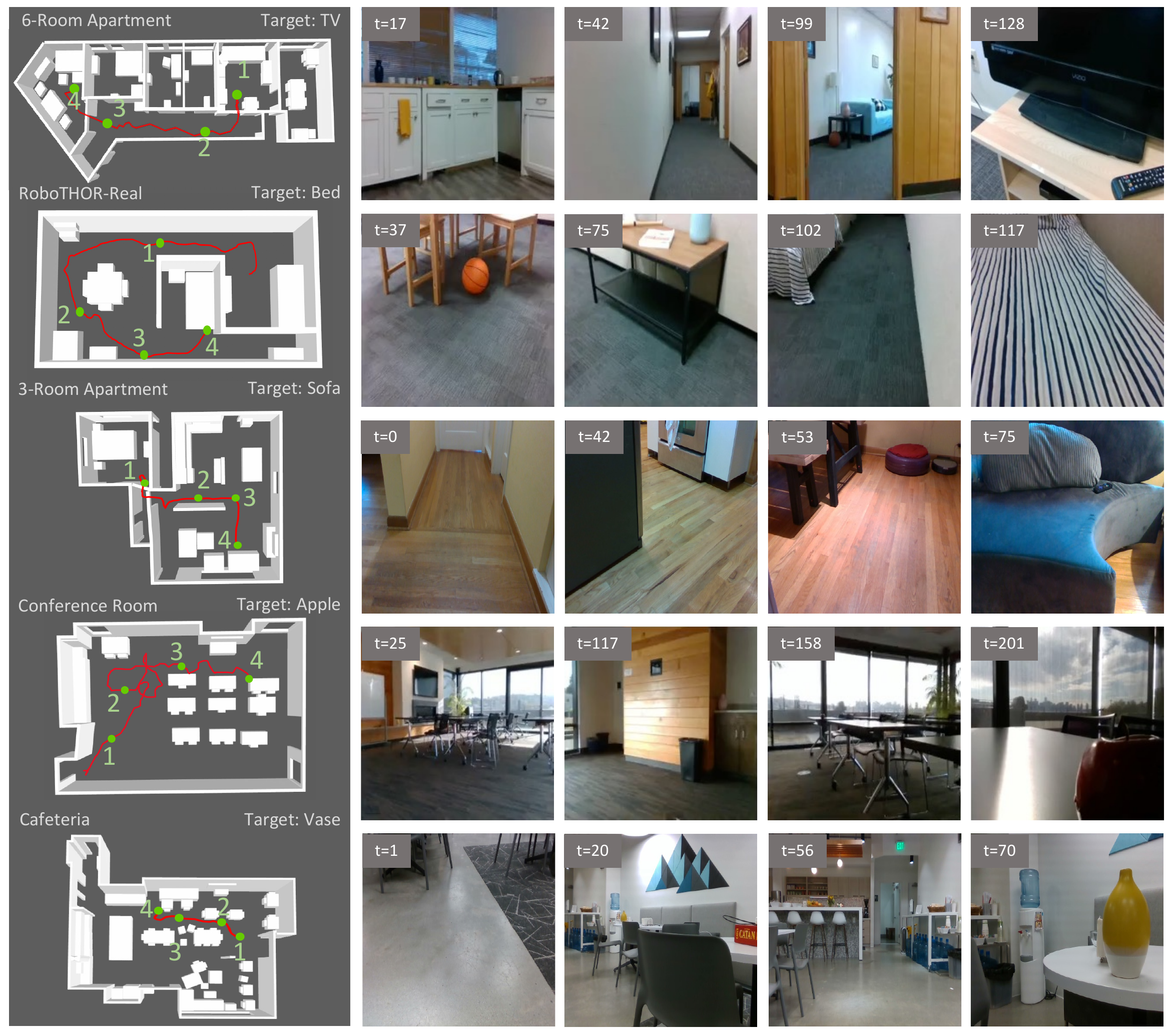}
    \caption{Qualitative results. These demonstrate the ability of \methodName models to navigate to their desired object. The top-down map is for visualization purposes only and is an approximation of the path taken by the agent. }
    \label{fig:qualitative-result}
    \vspace{-2em}
\end{figure}

\subsection{Qualitative Analysis}

Fig.~\ref{fig:qualitative-result} illustrates exemplary trajectories from each test environment with a few ego-centric RGB images that is the agent's only input. The trajectories show meaningfully different behavior for large vs. small objects (bed, sofa, and TV vs. apple and vase).

For large objects that don't change location drastically (\eg row 1), the agent seems to initially localize itself using known landmarks that appear in the scan (similar object categories, for instance) and then demonstrate efficient motion towards the room which contains the target object. We observed during our trials that often, when an agent navigating toward a large target loses its way, it would double back to a familiar large object and then restart direct progress. Note that the agent has no ground truth localization and must rely on its observations and its recollection of environment layout and object presence. 

In contrast to big objects, for smaller items, the agent needs to explore efficiently to find the target. For instance, in the fourth row of Fig.~\ref{fig:qualitative-result}, the agent searches for a small object that does not appear in the scan but is placed randomly in training rooms. Though again, it has no map, ground truth localization, or additional memory support, it demonstrates true exploration and high coverage of the possible area, ultimately achieving success (more qualitative results are provided in the supplementary videos).

\subsection{Quantitative Analysis of The Environments}
Results presented in Fig.~\ref{fig:four_env_results} span a wide range of space usage, layout, and complexity as quantified in Table~\ref{table:environment-analysis} to better demonstrate the power of \methodName. The conference room and cafeteria are large open spaces. The three living spaces require moving from room to room to locate objects. The 6-room apartment for instance, while most comparable in floor area to the conference room, is a long and narrow layout that requires hallway traversal for nearly every room transition. \methodName is most helpful in these environments over the baseline but makes a significant improvement in all layouts and semantic types of space.

\begin{table}[tp]
\footnotesize
\setlength{\tabcolsep}{2pt}
	\centering
\hfill
\begin{tabular}{l|ccccc}
Environment  & Area& Longest Path  & \# Rooms & \# Objects & \# Scanned \tabularnewline
  & (m$^2$) & (m) &  &  &Objects\tabularnewline
\toprule 
RoboTHOR-Real & 34.5 & 8.1 & 4 & 51 & 14 \tabularnewline
6-Room Apartment & 104.4 & 21.8 & 6 & 189 & 57 \tabularnewline
3-Room Apartment & 65.4 & 8.2 & 3 & 105 & 26 \tabularnewline
Conference Room & 98.3 & 10.0 & 1 & 48  & 32 \tabularnewline
Cafeteria & 133.2 & 18.8 & 1 & 252 & 67 \tabularnewline
\bottomrule

	\end{tabular}
\hfill 
	\vspace{0cm}
	\caption{The test real environments have a wide variety of layouts, usages, space, and object density. For visual layouts, see Fig.~\ref{fig:scan-sample}. }
	\label{table:environment-analysis}
 \vspace{-2em}
\end{table}

\section{Conclusion}
\label{sec:discussion}

In this paper, we introduced \methodName, a simple yet effective approach for training performant agents that are robust to the unpredictable nature of the real world. We demonstrated the capabilities of \methodName in five diverse environments and showed significant improvements in sim-to-real performance. Our environment-conditioned procedurally generated scenes are fully interactable, and we believe that future work will continue to explore its capabilities.

\mypara{Acknowledgements}
We would like to thank Klemen Kotar, Winson Han, Eli VanderBilt, Ian Grunfeld, and Alvaro Herrasti for assistance and discussions.

{\small
\bibliographystyle{ieee_fullname}
\bibliography{egbib}
}

\appendix

\section{Implementation Details}
For all experiments, we use the same architectures and process from EmbCLIP~\cite{khandelwal2022simple} and adopt the same hyperparameters as ProcTHOR~\cite{deitke2022procthor}. The 3x224x224 RGB images are processed with a frozen CLIP-ResNet-50 architecture \cite{he2016deep,radford2021learning}. This visual embedding is compressed with a 2-layer CNN, concatenated with a goal object type embedding, and compressed with a 2-layer CNN. This is flattened and combined with an embedding of the previous action, then passed through a single-layer GRU~\cite{chung2014empirical} policy with a hidden belief state of size 512. An actor and critic are used to generate the next action probability distribution and current state value estimates, respectively. The agent's next action is sampled from the actor distribution.

The following updates from \cite{deitke2022procthor} are made for policy and goal encoder fine-tuning:

\begin{enumerate}
\itemsep0em 
    \item Learning rate is lowered to 3e-5 (10x lower than that of ProcTHOR~\cite{deitke2022procthor}).
    \item A small penalty of -0.05 is assessed if the agent runs into objects.
    \item If the agent is about to run into an object, it will randomly move and rotate in small increments to coarsely emulate unmodeled and unintended physical environment interactions.
    \item The neck actions are limited to looking 30$^\circ$  above and below the horizon, as on our physical platforms.
    \item The horizontal field of view for a fixed aspect ratio is randomized by episode (uniformly sampled in 0.2$^{\circ}$ increments between 48$^{\circ}$ and 65$^{\circ}$), and the vertical field of view/aspect ratio is modified to more closely resemble the Intel RealSense D435.

\end{enumerate}
We use multi-node training on 3 or 4 (depending on the environment size) AWS g4dn.12xlarge machines with 16 processes per machine.

 The Habitat baseline used in Table~1 is trained on the 80 HM3D \cite{Ramakrishnan2021HabitatMatterport3D} set training scenes used for the 2022 Habitat challenge \cite{habitatchallenge2022} using 80 processes and 8 A100 GPUs. The model trained for 200M steps and we used the model which achieved the best performance on a validation set of 200 episodes. It was trained with the same updated field of view as every other model and baseline evaluated in this work.

\section{Failure Cases}

The most common failure case for \methodName models was semantic confusion; that is, not being able to recognize particular instances of objects or mistaking instances of other object categories for the target object. For example, a couch with a cover on it in the 6-room apartment was mistaken for a bed several times in the limited field of view of the agent and spare jugs for the water cooler in the cafeteria were mistaken as a vase. To generate the scenes, only six 3D models of vases were used. Thus, some semantic confusion is perhaps unsurprising, and \methodName with more visual diversity might be used to even greater effect.

\end{document}